\documentclass[conference]{IEEEtran}
\IEEEoverridecommandlockouts

\usepackage{amsmath,amssymb,amsfonts}
\usepackage{mathtools} 
\usepackage{algorithmic}
\usepackage{graphicx}
\usepackage{textcomp}
\usepackage{caption}
\usepackage{enumitem}
\usepackage{caption}
\usepackage{subcaption}
\usepackage[numbers]{natbib}
\def\BibTeX{{\rm B\kern-.05em{\sc i\kern-.025em b}\kern-.08em
    T\kern-.1667em\lower.7ex\hbox{E}\kern-.125emX}}

\usepackage{macro}
\usepackage{dirtytalk}

\usepackage{xcolor}
\definecolor{darkcerulean}{rgb}{0.03, 0.27, 0.49}
\definecolor{iris}{rgb}{0.35, 0.31, 0.81}
\definecolor{blue-violet}{rgb}{0.54, 0.17, 0.8}

\usepackage{hyperref}
\hypersetup{
     final=true,
     plainpages=false,
     pdfstartview=FitV,
     pdftoolbar=true,
     pdfmenubar=true,
     bookmarksopen=true,
     bookmarksnumbered=true,
     breaklinks=true,
     linktocpage=true,
     colorlinks=true,
     linkcolor=blue-violet, 
     urlcolor=iris, 
     citecolor=darkcerulean, 
     anchorcolor=black
}

\begin{document}

\title{Easing Optimization Paths: a Circuit Perspective
\thanks{\textsuperscript{*}Equal contribution.}
}

\author{
\IEEEauthorblockN{Ambroise Odonnat\textsuperscript{*}}
\IEEEauthorblockA{\textit{Noah’s Ark Lab, Inria}\\
\href{mailto:ambroise.odonnat@gmail.com}{ambroise.odonnat@gmail.com}}
\and
\IEEEauthorblockN{Wassim Bouaziz\textsuperscript{*}}
\IEEEauthorblockA{\textit{FAIR, Meta AI} \\
\href{mailto:wesbz@meta.com}{wesbz@meta.com}}
\and
\IEEEauthorblockN{Vivien Cabannes}
\IEEEauthorblockA{\textit{FAIR, Meta AI} \\
\href{mailto:vivc@meta.com}{vivc@meta.com}}
}

\maketitle

\begin{abstract}
    Gradient descent is the method of choice for training large artificial intelligence systems.
    As these systems become larger, a better understanding of the mechanisms behind gradient training would allow us to alleviate compute costs and help steer these systems away from harmful behaviors. 
    To that end, we suggest utilizing the circuit perspective brought forward by mechanistic interpretability.
    After laying out our intuition, we illustrate how it enables us to design a curriculum for efficient learning in a controlled setting. The code is available at \url{https://github.com/facebookresearch/pal}.

\end{abstract}


\section{Introduction}
\label{sec:intro}

Deep neural networks have attracted a lot of attention for their great empirical successes in many applications such as image classification~\cite{krizhevsky2012imagenet}, natural language processing~\cite{lample2019crosslingual}, protein folding prediction~\cite{jumper2021alphafold}, or playing chess~\cite{silver2016mastering}.
Recently, large language models became an emerging technology with worldwide use \cite{openai2024gpt4technicalreport,dubey2024llama3herdmodels}.
As the scaling of these models keeps increasing, the cost of their training becomes prohibitive.
This motivates studies regarding their training dynamics to minimize the cost per amount of learned intelligence.
In this introductory paper, we suggest that thinking in terms of circuits could provide valuable insights into this process.
This section introduces what circuits are, the setup to ground our thoughts, and a summary of our contributions.

\begin{figure}[!t]
    \centering
    \begin{subfigure}{0.5\linewidth}
        \includegraphics[width=\linewidth]{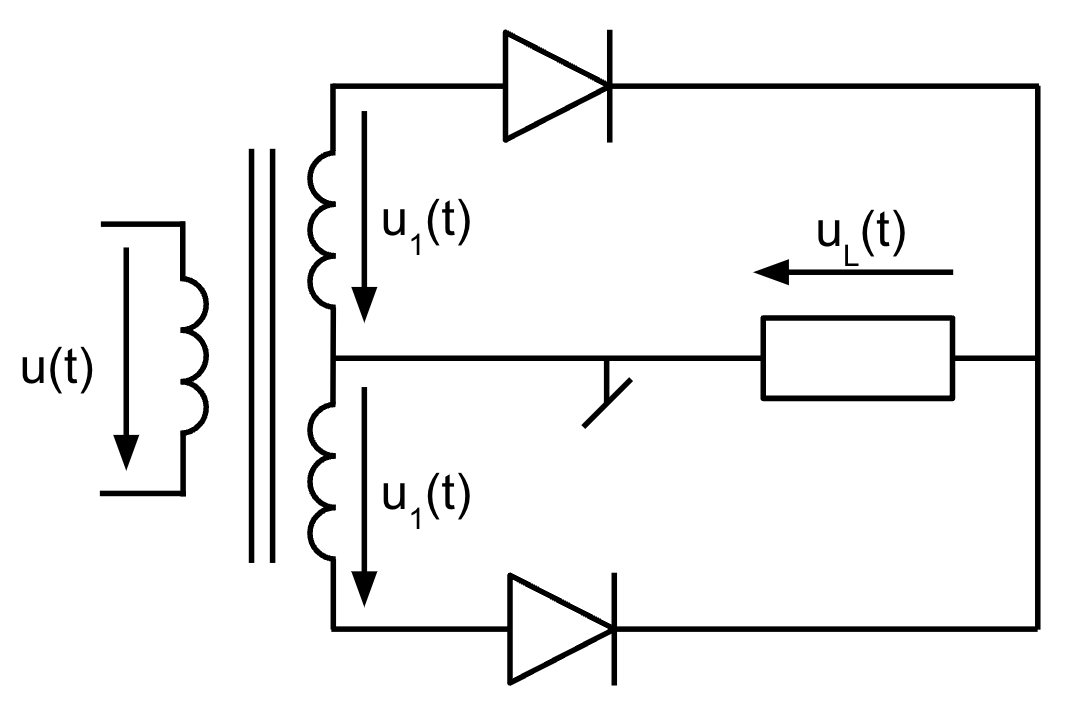}
        \caption{Electrical circuit.}
    \end{subfigure}
    \qquad
    \begin{subfigure}{0.3\linewidth}
        \includegraphics[width=\linewidth]{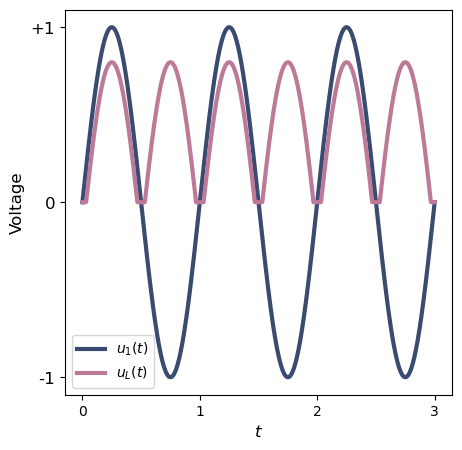}
        \caption{Response curve.}
    \end{subfigure}
    \begin{subfigure}{.9\linewidth}
        \includegraphics[width=\linewidth]{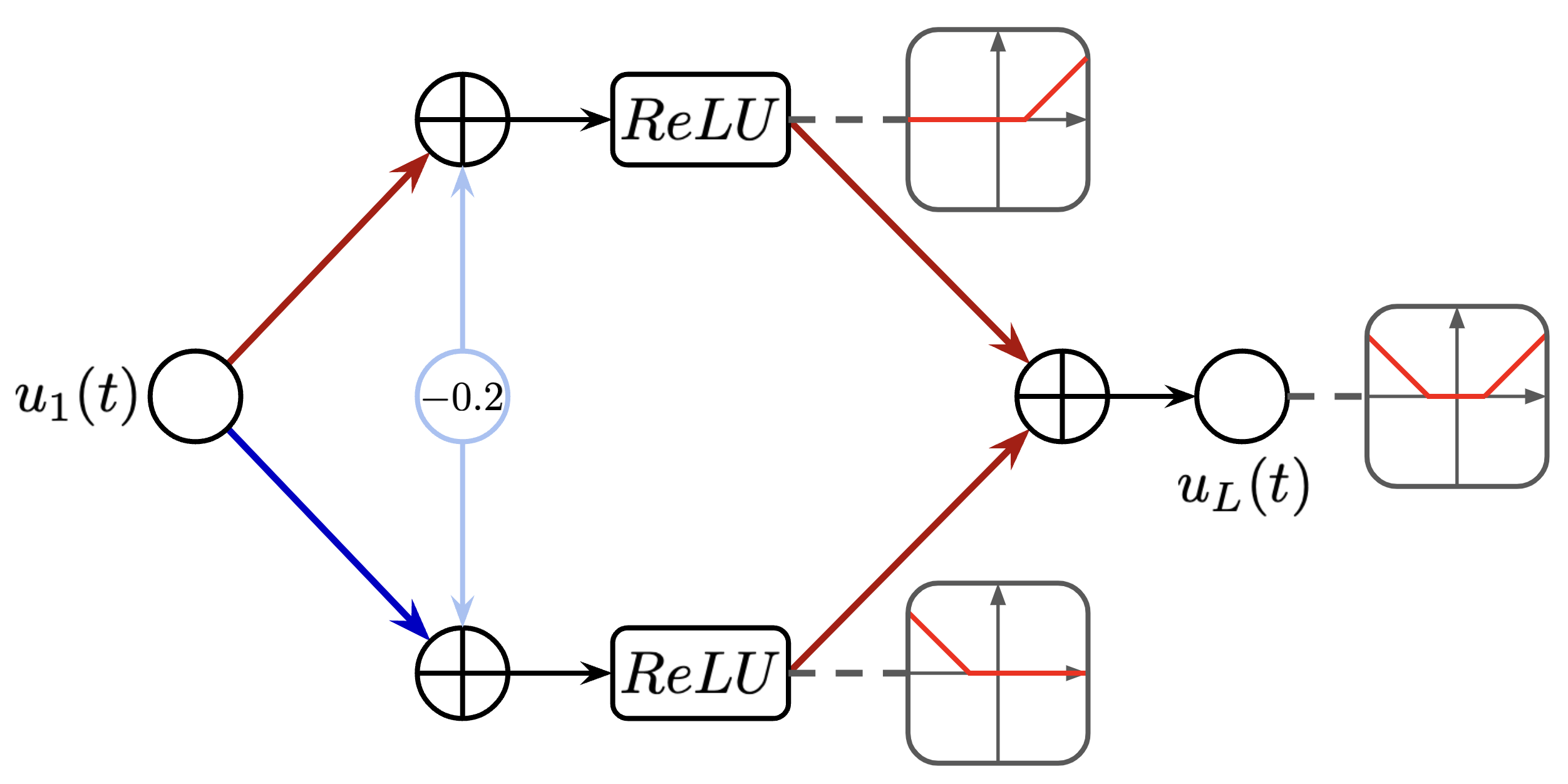}
        \caption{Neural network implementation.}
    \end{subfigure}
    \caption{Analogy between neural networks and electrical circuits, different components routing the electric/information flows.
    A center-tapped full wave rectifier can be implemented as a 2-layer neural network with ReLU activations. 
    Red and blue arrows represent respectfully +1 and -1 weights. The light blue represents the bias term.}
    \label{fig:nn_as_electrical_circuit}
\end{figure}

\paragraph{Circuits}
Artificial neural networks were inspired by the human brain, which is seen as a complex electrical circuit of interconnected neurons~\cite{rosenblatt1958perceptron}, where the flow of information is controlled by rectifiers~\cite{hahnloser2000digital}.
This is notably reflected in the naming of the most used activation layer, ReLU, for Rectified Linear Unit~\cite{fukushima1969visual}. 
Recently, the notion of \emph{circuit} has gained renewed attention in the field of mechanistic interpretability which attempts to reverse-engineer deep neural networks for greater understanding and reliability.
While the term circuit is polysemous, in this recent context, a circuit describes a pathway in a neural network that transforms some inputs into a given output~\cite{elhage2021mathematical}. 
A neural network can be thought of as a superposition of circuits and a circuit itself can be decomposed into sub-circuits. 
Defining circuits acting on some inputs can be somewhat subjective, aiming to capture semantically coherent units of calculation inside a network, such as algorithmic circuits, or memorization pathways~\cite{henighan2023superposition,nanda2023progressmeasuresgrokkingmechanistic}.

Thinking in terms of circuits opens new perspectives to comprehend neural networks, and information flowing through them.
The crux of this work is to study the dynamics of circuits along the training of neural networks.

\paragraph{Setup}
To carefully study the dynamics of circuits during training, and understand how the model utilizes these circuits, we focus on a simple task: the sparse modular addition problem.
Inputs are sequences of $T$ tokens in $\bF_p = \bZ / p\bZ \simeq [p]$, i.e., the ring of integers modulo $p$, and targets are the sum of the first $k$ terms. 
Formally,
\begin{equation*}
\begin{cases}
    x &= (x_1, x_2, \ldots, x_T) \text{ with } x_t \in \bF_p, \\
    y &= \sum_{t\leq k} x_t.
\end{cases}
\end{equation*}
In other terms, input sequences $x$ live in $\bF_p^T$, where $T$ is the sequence length and $p$ is the vocabulary size.
In practice, we take $p \in \{2, 4\}$, $T\in\{8, 12\}$, and $k=5$.

\paragraph{Summary of contributions}
This paper utilizes the concept of circuits to provide valuable insights into how neural networks learn and optimize their performance.
Our contributions can be summarized as follows:

\begin{enumerate}[leftmargin=*]
    \item  We explain what circuits are and their usefulness in understanding the training behavior of neural networks: gradient descent reinforces useful circuits while pruning others.
    \item  We discuss how gradient descent fosters sub-circuits, helping to solve complex tasks by breaking them down into intermediate reasoning steps.
    \item We detail the hardness of finding circuits and how curriculum learning and data curation are useful in easing their discovery by enhancing useful sub-circuits.
\end{enumerate}

Overall, our work provides a new perspective on understanding the training dynamics of neural networks and demonstrates the potential benefits of thinking in terms of circuits for optimizing their performance.

\paragraph{Architecture}
While the concept of circuit dynamics is agnostic to the choice of neural networks, we focus on the Transformer architecture~\cite{vaswani2017attention} for its great empirical successes.
Specifically, we consider a one-layer transformer with cross-attention.
Given an input sequence $x$ of length $T$ with a vocabulary size $p$, our model performs the following steps:

\begin{itemize}[leftmargin=*]
    \item \textbf{Token embeddings}. Each token $x_t$ is mapped to a $d$-dimensional embedding via an embedding matrix $W_E \in \bR^{d \times p}$. This results in $z_t = W_{E, x_t} \in \bR^d$, where $W_{E,j}$ is the $j$-th row of $W_E$;
    \item \textbf{Positional embeddings}. A learnable positional embedding $p_t \in \bR^d$ is added to each token $z_t$ depending on its position in the sequence, which is followed by a root-mean-square (RMS) normalization layer. 
    It results in embeddings of the form $z_t \coloneq \operatorname{RMS}(z_t + p_t)$, 
    \item \textbf{Attention block}. Given a sequence $z \in \bR^{d \times T}$, a query and a value matrices $W_Q \in \bR^d, W_V \in \bR^{d \times d}$, our cross-attention computes
    $$ z_A \coloneq (W_Vz) \operatorname{softmax}(\frac{z^\top W_Q}{\sqrt{d}}) \in \bR^d.$$
    Since $z$ can be set to anything thanks to $W_E$ and $p$, we omit the key, hence removing the need for an extra linear transformation $W_K z$.
    This is again followed by an RMS normalization,
    $$\bar{z}_{A} \coloneq \operatorname{RMS}(z_A) \in \bR^d.$$
    \item \textbf{Feed-forward block}. Finally, a feed-forward block is applied, consisting of a two-layer MLP with GELU activation denoted by $\operatorname{GELU}(x) = x \phi(x)$, where $\phi$ is the Gaussian cumulative function.
    It is followed by a residual connection.
    The output of this layer reads
    $$z_O \coloneq \bar{z}_{A} + W_2^\top \operatorname{GELU}(W_1 \bar{z}_{A}) \in \bR^d,$$
    with $W_1, W_2 \in \R^{h\times d}$ for $h$ a hidden dimension typically set to $h = 4d$.
    \item \textbf{Unembedding}. After the last transformer layer, the embedding vector $z_O\in\bR^d$ is mapped back to the vocabulary space through an unembedding matrix $W_U \in \bR^{p \times d}$. 
    The network output is a probability distribution
    $$
        p(y|x; W_U, z_O) \coloneq \operatorname{softmax}(W_U z_O).
    $$
\end{itemize}

The output of the model is then fed into a cross-entropy loss to predict the right target $y$ from the input sequence $x$. 
In all our experiments, the model is trained by gradient descent with Adam optimizer~\cite{kingma15adam}. Except specified otherwise, the learning rate is $lr=10^{-3}$ and the embedding dimension is $d=32$. \\

\section{Is Learning about Pruning Circuits?}

This section discusses gradient descent as a way to foster useful circuits and prune the other ones.
It equally introduces useful visualization to understand this pruning mechanism in attention modules.

\begin{figure}[!t]
    \centering
    \begin{subfigure}{0.22\textwidth}
        \includegraphics[width=\linewidth]{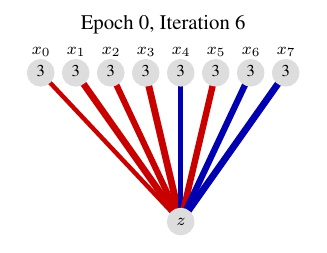}
    \end{subfigure}
    \qquad
    \begin{subfigure}{0.22\textwidth}
        \includegraphics[width=\linewidth]{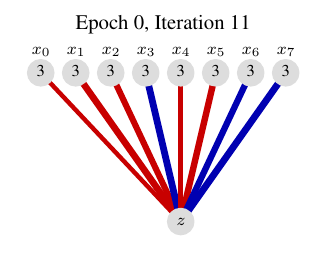}
    \end{subfigure}
    \begin{subfigure}{0.22\textwidth}
        \includegraphics[width=\linewidth]{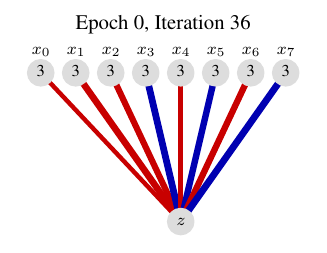}
        \caption{Typical initial update, increasing various connections.}
    \end{subfigure}
    \qquad 
    \begin{subfigure}{0.22\textwidth}
        \includegraphics[width=\linewidth]{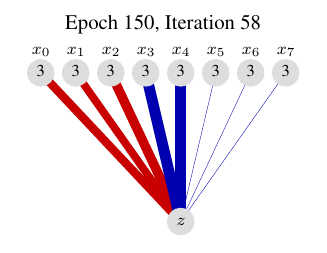}
        \caption{Final profile equalizing focus on non-spurious tokens.}
    \end{subfigure}
    \caption{
    Visualization of $(s_t) = \operatorname{softmax}((x_t + p\_t)^\top W_Q / \sqrt{d})$, the attention scores for a fixed sequence made up of $(x_t) = 3$, displayed at different points during training.
    The strength of the attention is visualized through the thickness of the arrows, while the color indicates the sign of the last updates: \textcolor{red!90!black}{red} for arrows that have just been thickened, \textcolor{blue!85!black}{blue} for those that have been thinned.
    }
    \label{fig:pruning}
\end{figure}

\paragraph{A Gradient Step May Enforce Many Circuits} 
To better comprehend the effect of an update, we visualize an attention layer using a bipartite graph structure where nodes represent both the input and output sequences and the edges are weighted according to the attention scores (the thicker the edge, the higher the attention). 
We illustrate the attention for a sequence with repeated entries on Figure~\ref{fig:pruning}.
To encompass the training dynamics, we store the attention maps along the iterations, and color in blue a weight that was just diminished, and in red one that was just increased.
At the beginning of training, the gradient updates sometimes reinforce attention on spurious tokens. 
This can be seen on the first three profiles 
of Figure \ref{fig:pruning}, where at the first epochs, among the 64 batches of size 32, some reinforce the 3 tokens among the 8 ones, i.e. $x_5$, $x_6$ or $x_7$, although they do not contribute to the sum $y$.
This is because some specific memorization circuits could be taken to memorize different batches.
Meanwhile, the updates tend to push the right attention scores, augmenting the focus on the first five tokens.

Our visualization helps build a better intuition of gradient descent.
During training, a gradient descent step updates the network weights proportionally to how much they \emph{locally} change the output. 
As circuits are spread across the network, several of them can be reinforced by one update.
A simple thought experiment can illustrate this mechanism.
Consider an input defined as the concatenation $[x_1, x_2]$ of twice the same original input $x = x_1 = x_2$. 
A gradient update will reinforce both circuits to go from $x_1$ to $y$ and from $x_2$ to $y$, likely creating duplicates, due to the locality of the updates.

\paragraph{Many Steps Pruned Non-Invariant Circuits}
Figure~\ref{fig:pruning} illustrates how, as we iterate over the data, spurious and non-spurious features are reinforced by gradient descent.
The spurious part consists of the 3 tokens on the right, i.e. $x_5$, $x_6$, and $x_7$, that do not participate in the definition of $y$.
The evolution of the corresponding attention weights is non-monotonic, going from blue to red and vice-versa.
On the contrary, the non-spurious part tends to be redder, ultimately becoming dominant, while the spurious circuits get pruned in the last profile.

We hypothesize that the phenomenon in Figure \ref{fig:pruning} is characteristic of gradient descent in neural networks.
As we iterate over data, gradient updates randomly increase or diminish connections in spurious circuits, while non-spurious connections tend to be increased more thoroughly.
As we iterate over pairs $(x, y)$, the circuits that invariantly help predict the right $y$ for all $x$ become dominant, pruning the other ones.

\begin{figure}[!t]
    \centering
    \includegraphics[width=\linewidth]{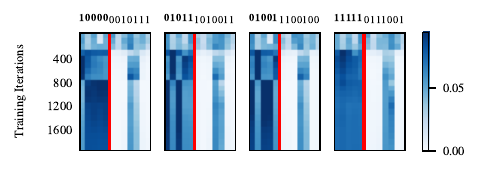}
    \caption{
    Evolution of the attention weights through gradient descent.
    Each line corresponds to a training iteration and each row corresponds to an entry $x_t$ of the input sequence $x$.
    The darker, the higher the attention weight.
    Ultimately, the transformer learns to focus solely on the first $k=5$ input tokens, which are the ones defining the output $y$, indicated by the \textcolor{red}{red} vertical line.
    More exactly, it focuses on the $0$ among these tokens, before counting them and deducing the number of $1$ to make its final prediction.
    }
    \label{fig:attention_training}
\end{figure}

\begin{figure}
    \centering
    \begin{subfigure}[t]{0.46\linewidth}
        \includegraphics[width=\linewidth]{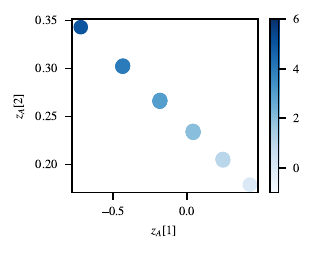}
        \caption{Pretraining with $p=2$.}
    \end{subfigure}
    \quad 
    \begin{subfigure}[t]{0.48\linewidth}
        \includegraphics[width=\linewidth]{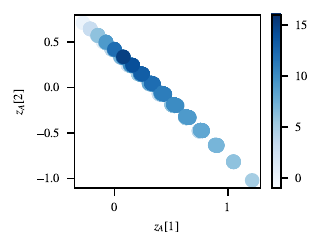}
        \caption{Finetuning with $p=4$.}
    \end{subfigure}
    \caption{Representation on the plan of the $d=2$ dimensional embeddings $z_A$ obtained after the attention module (see Section~\ref{sec:intro}). Colors represent the sum of the $k$ first tokens. \textbf{Left}: after pretraining with $p=2$, we observe the emergence of equivalence classes modulo $6$. After finetuning with $p=4$, equivalence classes modulo $16$ appear.}
    \label{fig:clusters_values}
\end{figure}

\paragraph{Learning Sub-Circuits}
When analyzing the circuit learned when $x_t\in\bF_p$ with $p=2$, we observe that the network first solves for $z = \sum_{t\leq k} x_t [k+1]$, before solving for $y = z [p]$.
This can be seen partially from Figure \ref{fig:attention_training}.
The attention focuses on the number of zeros among the $k$ non-spurious tokens, leading to 
\begin{align*}
    z_A
    &= \sum_{i < k} \ind{x_i = 0} (W_V W_{E,0} + W_Vp_i) 
    \\&= \sum_{i < k} \ind{x_i = 0} W_V W_{E,0} 
    \simeq k - \sum_{i<k} x_i [k+1],
\end{align*}
where the second equality is due to the structure of the learned position embedding and value matrix.
Figure~\ref{fig:clusters_values} shows the learned representations $z_{A}$ coming out of the attention block. Learning to solve our task for $p=2$ and $k=5$, the model learns to map the sequences into 6 clusters based on the sum modulo 6 of the first $k$ tokens of the sequence.
Similarly, finetuning from that model on $p=4$ data, the observed structure strengthens further and the embeddings are gathered in 16 different clusters.
Such a structure could not emerge from $p=4$ only on the considered model size ($d=2$).

In other terms, the network solves our task after splitting it into sub-tasks that involve intermediate reasoning steps.
This is reminiscent of how, in image processing, convolutional neural networks (CNNs) first extract low-level features, such as texture, and extract more and more fine-grained ones with increasing depth~\cite{rivas2023gabor}.
In this case, the first CNN layers resemble Gabor filters~\cite{gabor1946filter} which is similar to how human vision works.
The discovery of sub-circuits is not surprising.
During the forward pass, the information processed at a given layer is the result of sub-circuits in previous layers.
During the backward pass, gradient descent will reinforce sub-circuits in these previous layers that provide useful signals to build correct predictions.

\begin{figure}[!t]
    \centering
    \includegraphics[width=.7\linewidth]{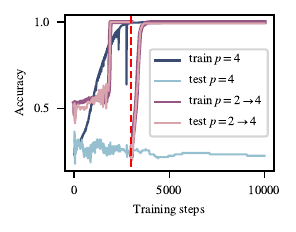}
    \caption{Evolution of the train and test accuracy along the training iterations. $p = 4$ corresponds to the model trained from scratch with $p=4$ and $p=2 \rightarrow 4$ is the model first pretrained with $p=2$ and then finetuned with $p=4$. The \textcolor{red}{red} dashed line indicates the iteration at which we switch from the pretraining to the finetuning.}
    \label{fig:comp_finetune_scratch}
\end{figure}

\section{Easing Optimization Paths}
The intuition laid out in the previous section could be of great interest for several reasons.
First, formalizing it into mathematical statements may help derive useful theorems.
Second, it may help discover better optimization schemes.
To that end, this section provides a compelling experiment in our controlled setup.
It showcases how carefully chosen curriculum learning can instill the right circuits to solve a needle-in-a-haystack problem.

When trying to solve the modular addition problem with $p=4$, $T=12$, and $k=5$ for $n=2048$ training data, we need to find a circuit specified for $2^{11}$ points that generalize to $12^{4} > 2^{14}$ points.
This is akin to finding a needle in a haystack, and as shown by Figure \ref{fig:comp_finetune_scratch}, training from scratch leads to networks that memorize their training sets but do not generalize to our testing set (blueish curves).
However, if we first train the network with parity data, i.e. $x_t \in \bF_p$ with $p=2$, we have seen that the first attention layer will implement an addition modulo $k+1$.
When initializing a network with such an already implemented sub-circuit, and then training with $p=4$, gradient descent easily finds a circuit that can generalize to the $12^4$ data (pinkish curves).
In practice, we train the network with a dataset where $p=2$ for 3000 epochs in full batch, before switching the training data to $p=4$ and continuing the training for another 7000 epochs.
This explains the discontinuity (horizontal dashed line) in Figure \ref{fig:comp_finetune_scratch}.

\begin{figure}[!t]
    \centering
    \includegraphics[width=.7\linewidth]{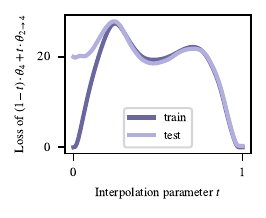}
    \caption{
    Loss profile of $f_{(1 - t)\cdot \theta_{4} + t\cdot \theta_{2\to 4}}$, where $\theta$ denotes the weights of the network $f_\theta$, $\theta_4$ are the weights learned from scratch, and $\theta_{2\to 4}$ are the weights learned from curriculum technique of Figure \ref{fig:comp_finetune_scratch}.
    It shows that to go from $\theta_4$ to $\theta_{2\to 4}$ has to cross a high potential barrier, making it hard to escape bad local minima.
    }
    \label{fig:interp_loss}
\end{figure}

Figure~\ref{fig:interp_loss} illustrates the potential barrier to going from the memorizing solution learned from scratch and the generalizing ones learned with curriculum for our problem with $p=4$.
Loss profile of $f_{t\cdot \theta_{4} + (1 - t)\cdot \theta_{2\to 4}}$, where $\theta$ denotes the weight of the network $f_\theta$, $\theta_4$ are the weights learned from scratch, and $\theta_{2\to 4}$ are the weights learned with the curriculum technique of Figure \ref{fig:comp_finetune_scratch}.
The high loss barrier separating the two models explains why the model trained from scratch fails at leaving its local minimum and learning a general solution.

Once again, the observations made on our controlled problem are meaningful more generically.
For complicated problems, finding the right circuits can be extremely hard, and require a large amount of data.
In these cases, networks with high capacity can simply store their training data within disjoint memorization pathways, without generalizing to unseen data.
However, in many cases, there exist solutions based on sub-circuits that are easier to learn with proxy tasks.
In particular, curriculum learning may help learn modular components that can later be used to learn more complicated tasks \citep[see][for related observations]{abbe2023provable,cabannes2024iteration}.
This may explain the practical usefulness of curriculum learning \cite{odonnat2024diversity, cascante2021curriculum, ionescu2016howhard, platonios2019curriculum}, as well as the usefulness of careful data curation to enhance the right circuits \cite{dubey2024llama3herdmodels}.

\begin{figure}[!t]
    \centering
        \includegraphics[width=\linewidth]{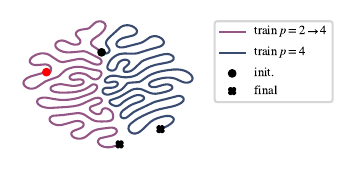}
        \caption{Low-dimensional t-SNE representation of the transformer's $10000$ parameters during the gradient descent. The black circle represents the initial point and the black crosses represent the end point at the end of training. For the model $p = 2 \rightarrow 4$ that is first pre-trained with $p=2$ and then finetuned with $p=4$, the \textcolor{red}{red} circle represents the switch between pretraining and finetuning.}
    \label{fig:enter-label}
\end{figure}

\section{Conclusion}
In this paper, we propose a new perspective on understanding the training dynamics of neural networks through the lens of circuits.
It lays out intuition regarding the benefits of this perspective and showcases how it helps design curricula to easily expose useful circuits and optimize network performance.
We believe that this perspective has the potential to advance the field of deep learning both in terms of practical results and theoretical understanding. 

Future work could use a circuit perspective to study the interplay between gradient norms, loss spikes, and training stability~\cite{zhang2024transformersneedadamhessian, ilbert2024samformer} or the impact of activation sparsity on models' performance~\cite{li2023thelazy,mirzadeh2024relu}. Our work could enable researchers to first elucidate them on a small scale with the benefits of an easier visualization before tackling them on a larger scale.

\bibliography{reference}

\begin{thebibliography}{10}

\bibitem{abbe2023provable}
Emmanuel Abbe, Elisabetta Cornacchia, and Aryo Lotfi.
\newblock Provable advantage of curriculum learning on parity targets with mixed inputs.
\newblock In {\em Advances in Neural Information Processing Systems}, volume~36, pages 24291--24321. Curran Associates, Inc., 2023.

\bibitem{cabannes2024iteration}
Vivien Cabannes, Charles Arnal, Wassim Bouaziz, Alice Yang, Francois Charton, and Julia Kempe.
\newblock Iteration head: A mechanistic study of chain-of-thought, 2024.

\bibitem{cascante2021curriculum}
Paola Cascante-Bonilla, Fuwen Tan, Yanjun Qi, and Vicente Ordonez.
\newblock Curriculum labeling: Revisiting pseudo-labeling for semi-supervised learning.
\newblock In {\em Proceedings of the AAAI conference on artificial intelligence}, volume~35, pages 6912--6920, 2021.

\bibitem{lample2019crosslingual}
Alexis Conneau and Guillaume Lample.
\newblock Cross-lingual language model pretraining.
\newblock In {\em Advances in Neural Information Processing Systems}, volume~32. Curran Associates, Inc., 2019.

\bibitem{elhage2021mathematical}
Nelson Elhage, Neel Nanda, Catherine Olsson, Tom Henighan, Nicholas Joseph, Ben Mann, and et~al.
\newblock A mathematical framework for transformer circuits.
\newblock {\em Transformer Circuits Thread}, 2021.
\newblock https://transformer-circuits.pub/2021/framework/index.html.

\bibitem{fukushima1969visual}
Kunihiko Fukushima.
\newblock Visual feature extraction by a multilayered network of analog threshold elements.
\newblock {\em IEEE Transactions on Systems Science and Cybernetics}, 5(4):322--333, 1969.

\bibitem{gabor1946filter}
D.~Gabor.
\newblock Theory of communication.
\newblock {\em J. Inst. Electr. Engineering}, pages 429--457, 1946.

\bibitem{hahnloser2000digital}
Richard H.~R. Hahnloser, Rahul Sarpeshkar, Misha~A. Mahowald, Rodney~J. Douglas, and H.~Sebastian Seung.
\newblock Digital selection and analogue amplification coexist in a cortex-inspired silicon circuit.
\newblock {\em Nature}, 405(6789):947--951, Jun 2000.

\bibitem{henighan2023superposition}
Tom Henighan, Shan Carter, Tristan Hume, Nelson Elhage, Robert Lasenby, Stanislav Fort, Nicholas Schiefer, and Olah Christopher.
\newblock Superposition, memorization, and double descent.
\newblock {\em Transformer Circuits Thread}, 2023.
\newblock https://transformer-circuits.pub/2023/toy-double-descent/index.html.

\bibitem{ilbert2024samformer}
Romain Ilbert, Ambroise Odonnat, Vasilii Feofanov, Aladin Virmaux, Giuseppe Paolo, Themis Palpanas, and Ievgen Redko.
\newblock {SAM}former: Unlocking the potential of transformers in time series forecasting with sharpness-aware minimization and channel-wise attention.
\newblock In {\em ICML}, 2024.

\bibitem{ionescu2016howhard}
Radu~Tudor Ionescu, Bogdan Alexe, Marius Leordeanu, Marius Popescu, Dim Papadopoulos, and Vittorio Ferrari.
\newblock How hard can it be? estimating the difficulty of visual search in an image.
\newblock In {\em Proceedings of CVPR}, pages 2157--2166, 06 2016.

\bibitem{jumper2021alphafold}
John~M. Jumper, Richard Evans, Alexander Pritzel, Tim Green, Michael Figurnov, Olaf Ronneberger, Kathryn Tunyasuvunakool, Russ Bates, Augustin Ž{\'i}dek, Anna Potapenko, and et~al.
\newblock Highly accurate protein structure prediction with alphafold.
\newblock {\em Nature}, 596:583 -- 589, 2021.

\bibitem{kingma15adam}
Diederik Kingma and Jimmy Ba.
\newblock Adam: A method for stochastic optimization.
\newblock In {\em International Conference on Learning Representations (ICLR)}, San Diega, CA, USA, 2015.

\bibitem{krizhevsky2012imagenet}
Alex Krizhevsky, Ilya Sutskever, and Geoffrey~E Hinton.
\newblock Imagenet classification with deep convolutional neural networks.
\newblock In {\em Advances in Neural Information Processing Systems}, volume~25. Curran Associates, Inc., 2012.

\bibitem{li2023thelazy}
Zonglin Li, Chong You, and Srinadh~Bhojanapalli et~al.
\newblock The lazy neuron phenomenon: On emergence of activation sparsity in transformers.
\newblock In {\em ICLR}, 2023.

\bibitem{dubey2024llama3herdmodels}
Meta.
\newblock The llama 3 herd of models, 2024.

\bibitem{mirzadeh2024relu}
Seyed~Iman Mirzadeh, Keivan Alizadeh-Vahid, Sachin Mehta, and Carlo~C del Mundo~et al.
\newblock Re{LU} strikes back: Exploiting activation sparsity in large language models.
\newblock In {\em ICLR}, 2024.

\bibitem{nanda2023progressmeasuresgrokkingmechanistic}
Neel Nanda, Lawrence Chan, Tom Lieberum, Jess Smith, and Jacob Steinhardt.
\newblock Progress measures for grokking via mechanistic interpretability, 2023.

\bibitem{odonnat2024diversity}
Ambroise Odonnat, Vasilii Feofanov, and Ievgen Redko.
\newblock Leveraging ensemble diversity for robust self-training in the presence of sample selection bias.
\newblock In Sanjoy Dasgupta, Stephan Mandt, and Yingzhen Li, editors, {\em Proceedings of The 27th International Conference on Artificial Intelligence and Statistics}, volume 238 of {\em Proceedings of Machine Learning Research}, pages 595--603. PMLR, 02--04 May 2024.

\bibitem{openai2024gpt4technicalreport}
OpenAI.
\newblock Gpt-4 technical report, 2024.

\bibitem{platonios2019curriculum}
Emmanouil~Antonios Platanios, Otilia Stretcu, Graham Neubig, Barnabás Póczos, and Tom~M. Mitchell.
\newblock Competence-based curriculum learning for neural machine translation.
\newblock In Jill Burstein, Christy Doran, and Thamar Solorio, editors, {\em NAACL-HLT (1)}, pages 1162--1172. Association for Computational Linguistics, 2019.

\bibitem{rivas2023gabor}
Pablo Rivas and Mehang Rai.
\newblock Gabor filters as initializers for convolutional neural networks: A study on inductive bias and performance on image classification.
\newblock In {\em LatinX in AI Workshop at ICML 2023 (Regular Deadline)}, 2023.

\bibitem{rosenblatt1958perceptron}
F.~Rosenblatt.
\newblock {The perceptron: A probabilistic model for information storage and organization in the brain.}
\newblock {\em Psychological Review}, 65(6):386--408, 1958.

\bibitem{silver2016mastering}
David Silver, Aja Huang, Chris~J Maddison, Arthur Guez, Laurent Sifre, George Van Den~Driessche, Julian Schrittwieser, Ioannis Antonoglou, Veda Panneershelvam, Marc Lanctot, et~al.
\newblock Mastering the game of go with deep neural networks and tree search.
\newblock {\em Nature}, 529(7587):484--489, 2016.

\bibitem{vaswani2017attention}
Ashish Vaswani, Noam Shazeer, Niki Parmar, Jakob Uszkoreit, Llion Jones, Aidan~N Gomez, \L~ukasz Kaiser, and Illia Polosukhin.
\newblock Attention is all you need.
\newblock In {\em Advances in Neural Information Processing Systems}, volume~30. Curran Associates, Inc., 2017.

\bibitem{zhang2024transformersneedadamhessian}
Yushun Zhang, Congliang Chen, Tian Ding, Ziniu Li, Ruoyu Sun, and Zhi-Quan Luo.
\newblock Why transformers need adam: A hessian perspective.
\newblock In {\em NeurIPS}, 2024.

\end{thebibliography}
\bibliographystyle{plain}

\end{document}